\documentclass[letterpaper]{article} 
\usepackage{aaai24}  
\usepackage{times}  
\usepackage{helvet}  
\usepackage{courier}  
\usepackage[hyphens]{url}  
\usepackage{graphicx} 
\usepackage{natbib}  
\usepackage{caption} 
\usepackage{algorithm}
\usepackage{algorithmic}
\usepackage{xcolor}

\usepackage{newfloat}
\usepackage{listings}
\DeclareCaptionStyle{ruled}{labelfont=normalfont,labelsep=colon,strut=off} 
\floatstyle{ruled}
\newfloat{listing}{tb}{lst}{}
\floatname{listing}{Listing}
\title{Knowledge-Powered Recommendation for an Improved Diet Water Footprint}
\author{
    Saurav Joshi\textsuperscript{\rm 1},
    Filip Ilievski\textsuperscript{\rm 1,2},
    Jay Pujara\textsuperscript{\rm 1}
}

\affiliations {
    \textsuperscript{\rm 1}USC Information Sciences Institute, Marina del Rey, CA, USA\\
    \textsuperscript{\rm 2}Vrije Universiteit, Amsterdam, The Netherlands\\
    syjoshi@isi.edu, f.ilievski@vu.nl, jpujara@isi.edu
}

\usepackage{bibentry}

\begin{document}

\maketitle

\begin{abstract}
According to WWF, 1.1 billion people lack access to water, and 2.7 billion experience water scarcity at least one month a year. By 2025, two-thirds of the world's population may be facing water shortages. This highlights the urgency of managing water usage efficiently, especially in water-intensive sectors like food. This paper proposes a recommendation engine, powered by knowledge graphs, aiming to facilitate sustainable and healthy food consumption. The engine recommends ingredient substitutes in user recipes that improve nutritional value and reduce environmental impact, particularly water footprint. The system architecture includes source identification, information extraction, schema alignment, knowledge graph construction, and user interface development. The research offers a promising tool for promoting healthier eating habits and contributing to water conservation efforts.
\end{abstract}

\section{Introduction}

Water scarcity is a prevailing issue that poses significant environmental concerns \cite{rijsberman2006water}. This essential life-giving resource is critical not just for humans but also for the planet at large. Consequently, comprehending and regulating its consumption has become a crucial responsibility in the contemporary context. A key initiative in this realm involves the development of the metric known as the \textit{water footprint} which measures and raises consciousness about water usage \cite{hoekstra2009water}.

The global food sector is a significant contributor to climate change \cite{doi:10.1146/annurev-environ-020411-130608}, biodiversity loss and land-use change \cite{doi:10.1126/science.1111772}, and rampant exhaustion of freshwater resources \cite{https://doi.org/10.1029/2010GL044571}. A comprehensive study \cite{w12102696} indicates that the food industry is one of the most water-intensive sectors. To effectively manage water usage, numerous studies have been directed to establish dietary guidelines that advocate a more sustainable approach to food consumption with minimal environmental impacts \cite{springmann2018options,mekonnen2018effect}. However, despite the clear importance of healthy eating, many struggle with it due to busy lifestyles or lack of meal planning motivation \cite{10.1145/1943403.1943422}. Health-aware food recommender systems use both content-based approaches, derived from user and item attributes \cite{10.1007/978-3-319-02432-5_19}, and collaborative filtering, which bases predictions on historical community preferences \cite{10.1145/2750511.2750528}. In~\cite{8938875}, authors explore knowledge graphs as a solution for recommendation systems. Moreover, in recipe recommendations, these graphs view recipes as feature aggregates, enhancing prediction accuracy \cite{da2021knowledge}.


Leveraging these advancements, this paper proposes the development of a recommendation engine powered by knowledge graphs. It aims to help users navigate their diet planning efforts more sustainably by proposing substitutes for food ingredients in user's recipes that offer improved nutritional value and reduce the environmental impact regarding sustainability and water footprint. This concept stems from the ability of the knowledge graphs to offer context-aware recommendations, reveal hidden or implicit relationships, and facilitate informed decision-making; thereby serving as an effective tool for users to eat healthily and sustainably.

\section{System Architecture}
\begin{figure*}[t]
\centering
\includegraphics[width=\textwidth]{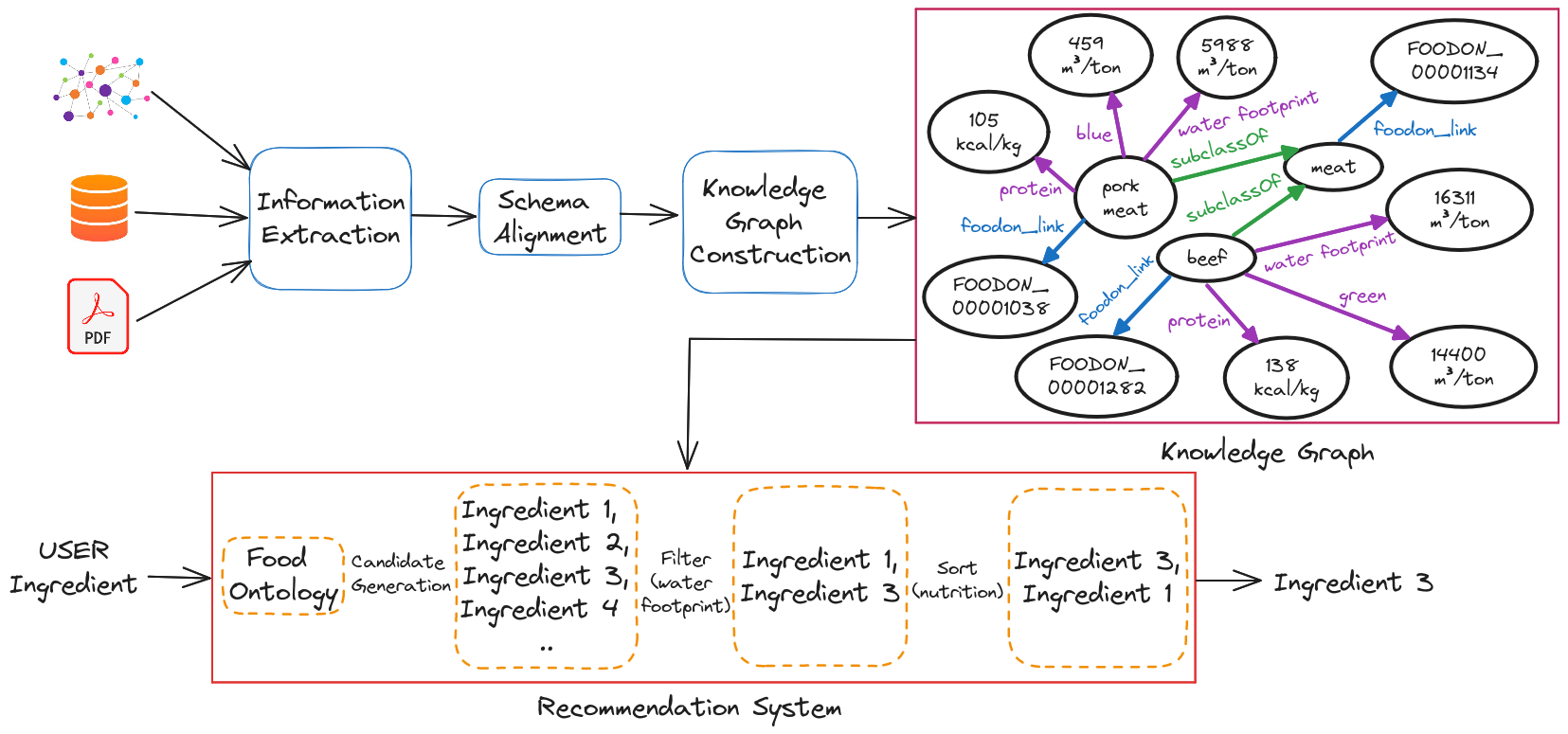} 
\caption{System Architecture consisting of five steps: Source Identification, Information Extraction, Schema Alignment, Knowledge Graph Construction, and User Interface Development.}
\label{fig:sys}
\end{figure*}

As seen in Figure \ref{fig:sys},
the system architecture comprises five sequential steps: Source Identification, Information Extraction, Schema Alignment, Knowledge Graph Construction, and User Interface Development.

\noindent \textbf{Source Identification.} Our research employs the FoodKG resource \cite{10.1007/978-3-030-30796-7_10}, a comprehensive database spanning over a million recipes, ingredients, and nutrients, encompassing approximately 67 million data points. We interpret and process information from three primary sources to lay the foundation for FoodKG: the Recipe1M dataset\cite{marin2021recipe1m+}, the USDA\footnote{\url{https://www.usda.gov/topics/food-and-nutrition}} database which catalogs the nutritional content of each ingredient, and FOODON which is an extensive food ontology system for ingredient organization\cite{dooley2018foodon}. Additionally, we incorporate complementary data on water footprints from academic journals and papers publicly available as PDF documents. 

\noindent \textbf{Information Extraction.} Our approach involves the application of the Knowledge Graph Toolkit (KGTK) \cite{ilievski2021kgtk}. We transform the information from the FoodKG, initially in the trig format, into nt triples format and, subsequently, to the KGTK format. Using Kypher~\cite{chalupsky2021creating} we successfully extract granular nutrient and FOODON data from the broader dataset. Afterward, we use the Semantic Scholar Open Research Corpus (S2ORC) PDF parser \cite{lo-etal-2020-s2orc} to extract information from academic articles.

\noindent \textbf{Schema Alignment.} We align ingredient data from three sources: nutritional content, the FOODON system, and water footprint information. First, we analyze ingredient names to extract main descriptors. This step simplifies complex names like 'vanilla-flavored soy yogurt' to 'soy yogurt'. Next, we turn graph relations into English sentences to compute sentence embeddings using Sentence Transformers. These embeddings serve as features for the model developed in the "Knowledge Graph Construction" phase.

\noindent \textbf{Knowledge Graph Construction.} We unify data from various sources using a common ingredient identifier. Addressing the challenge of missing entities in our knowledge base, we employ a neural network architecture: four dense layers followed by an output layer with a continuous ReLu activation function. Given an ingredient's embedding and its relation type, we predict the nutritional and water footprint value. It's optimized using the Adam optimizer and MSE loss. Our graph comprises of 20,778 nodes and 13 types of relationships, responding within an average latency of 1s.

\noindent \textbf{User Interface Development.} We design a user-centric interface via the Django framework. Users input their recipe ingredients, and via the Knowledge Graph we recommend alternatives with a lower water footprint, thus being more sustainable. We host this application on the AWS\footnote{\url{https://aws.amazon.com/}} platform.

\section{Demonstration}

The demonstration video showcases the efficacy of our recommendation engine in promoting sustainable eating by emphasizing the water footprint. Upon entering recipe ingredients like butter cream, whipped cream, and chocolate fudge cookie, the system first discerns the 'parent ingredient' through the food ontology hierarchy. It then chooses 'candidate ingredients' from a similar tier, excluding those with a larger water footprint. Subsequently, ingredients are ranked, prioritizing those with the least water consumption. For instance, when users aim to reduce fat, the engine suggests alternatives. After opting for a recommended ingredient, the system visualizes the water footprint difference alongside any coincidental nutritional changes. In this example, the total water footprint significantly diminishes from 20,135m³/ton to 5,215m³/ton. While the primary focus is on water conservation, it's noteworthy that the fat content also decreases from 30.32g to 11.84g, underscoring the holistic benefits of our system.

\section{Conclusion}

In this work, we introduced a recommendation tool centered on the concept of water footprint, utilizing a knowledge graph pipeline to advocate for sustainable eating habits. Our tool prioritizes ingredient substitutions that not only enhance nutritional value but also significantly reduce the water footprint of recipes. The aim is to directly influence individual food choices in favor of sustainability and water conservation. Looking ahead, our future endeavors will involve conducting user studies, expanding our ingredient database with a focus on water footprint data, and refining the recommendation algorithm. 

\section{Acknowledgments}

We thank Yuhua Wu and Shreya Padmanabhan for their invaluable contributions. This work was funded by the Defense Advanced Research Projects Agency with award HR00112220046.

\bibliography{aaai24}

\end{document}